\documentclass[letterpaper, 10 pt, conference]{ieeeconf}  % Comment this line out
\IEEEoverridecommandlockouts                              % This command is only
\usepackage{xcolor}
\usepackage{comment}
\usepackage{graphics} % for pdf, bitmapped graphics files
\usepackage{graphicx}
\usepackage{subcaption}
\usepackage{epsfig} % for postscript graphics files
\usepackage{amsmath} % assumes amsmath package installed
\usepackage{amssymb}  % assumes amsmath package installed
\usepackage{mathtools}
\usepackage{marginnote}
\usepackage{bm}
\usepackage{hyperref}
\usepackage{iitem}
\usepackage{float}
\usepackage{array}

\usepackage{booktabs}
\usepackage{svg}
\usepackage{balance}
\usepackage{caption}
\hypersetup{hidelinks}
\title{\LARGE \bf
Humanoid Rickshaw Pulling: Whole-Body Locomotion under \\ Coupled Wheeled Loads} % may change the name 

\author{
Yangzhi Yang$^{1}$,  Xiansheng Lin$^{1}$, Zhaoming Xie, and Xiaobin Xiong$^{1,\dagger}$\\
$^{1}$ Legged AI Lab, Shanghai Innovation Institute\\
$^{\dagger}$Corresponding to: xiaobin.xiong@sii.edu.cn, Video: \url{https://youtu.be/eqnAlQLjZF8}
}

\IEEEaftertitletext{%
\begin{minipage}{\textwidth}
    \vspace{-8mm}
    \centering
    \includegraphics[width=\textwidth]{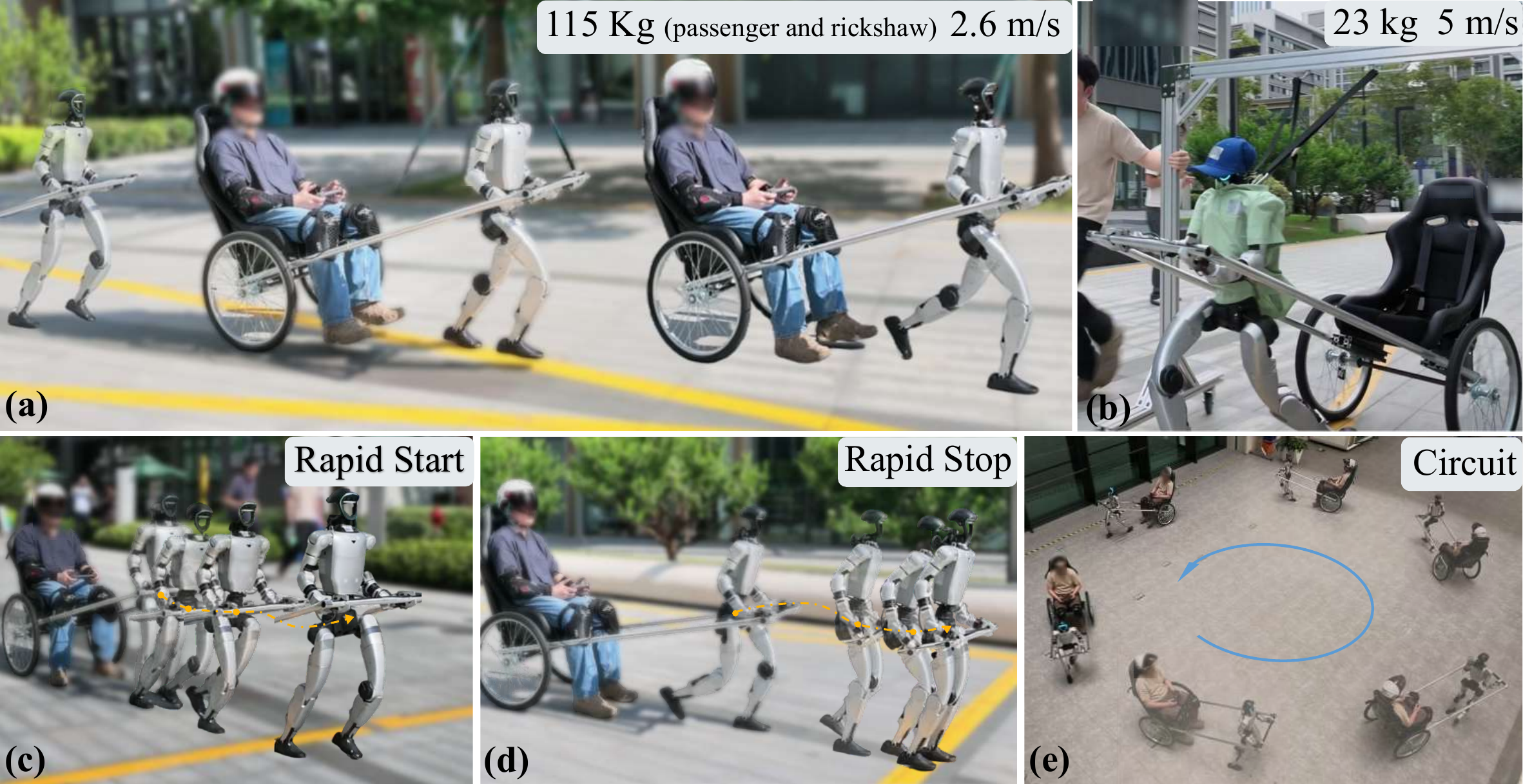}
    \captionof{figure}{\textbf{Overview.}
    (a) Our whole-body controller enables the humanoid to
    pull a 115~kg combined passenger-and-rickshaw load at 2.6~m/s.
    (b) High-speed pulling at 5~m/s with a 23~kg load.
    (c) Rapid start.
    (d) Rapid braking.
    (e) Circular pulling.}
    \label{fig:rickshaw_overview}
    \vspace{7pt}
\end{minipage}%
}

\begin{document}

 \newcommand{\block}[1]{\noindent{\textbf{#1}:}}
\newcommand{\fix}[1]{{\color{red} #1}}

\newcommand{\todo}[1]{{\color{yellow} #1}}

\newcommand{\todoYang}[1]{{\color{blue} #1}}

\newcommand{\todoLin}[1]{{\color{red} #1}}
\newcommand{\todoSong}[1]{{\color{red} #1}}

\newcommand{\todoXie}[1]{{\color{red} #1}}

\maketitle
\thispagestyle{empty}
\pagestyle{empty}

%%%%%%%%%%%%%%%%%%%%%%%%%%%%%%%%%%%%%%%%%%%%%%%%%%%%%%%%%%%%%%%%%%%%%%%%%%%%%%%%

%%% Dry run. Moodify it as you go. 
\begin{abstract}
Humanoid robots could transport payloads substantially heavier than themselves by pulling passive wheeled vehicles instead of carrying the load. This capability, however, creates a coupled locomotion problem: the robot must maintain persistent upper-body contact while adapting to unknown, configuration-dependent forces arising from the payload, vehicle, and terrain. We present a whole-body control framework for humanoid rickshaw pulling that tracks commanded vehicle motion while preserving balance and stable grasps under uncertain load dynamics. During training, a privileged teacher exploits vehicle states, interaction forces, and load properties. Its actions and latent are distilled into a history-conditioned student that implicitly infers coupled dynamics from proprioceptive responses, followed by reinforcement-learning fine-tuning. Comparisons with \emph{No History} and \emph{Only History} baselines show that the resulting policy achieves accurate vehicle tracking while reducing vehicle oscillation, torso tilt, and actuation cost. Behavioral analysis shows that Unitree G1 propels the rickshaw and generates gait-synchronized whole-body reactions that stabilize its lateral and roll motions. Moreover, pulling redistributes joint effort and yields a lower robot-normalized cost-of-transport proxy than unloaded walking over most tested load--speed conditions. On hardware, a single policy performs starting, sustained pulling, turning, and stopping with both rigid payloads and human passengers, handling a loaded rickshaw mass of up to 115~kg without load-specific retuning. These results demonstrate robust heavy-load transportation through coordinated and persistent humanoid--vehicle interaction.
\end{abstract}

%**Abstract**

\section{Introduction}
\label{sec:introduction}

Humanoid robots are expected to transport people, tools, and goods in human-centered environments. Directly carrying a payload places substantial demands on joint torques, structural strength, and balance. Pulling a passive wheeled vehicle offers an alternative: the wheels bear much of the payload weight, while the robot supplies propulsion and steering through sustained physical contact. This arrangement may enable transport of loads substantially heavier than the robot itself. We study \emph{humanoid rickshaw pulling} as a representative task in which whole-body locomotion must control a coupled wheeled load.

The central challenge is to regulate vehicle motion while maintaining humanoid balance and persistent bimanual contact. Forces transmitted through the handles couple the vehicle’s motion to the robot’s arms and torso. These forces depend on payload mass and distribution, wheel–ground interaction, and the relative robot–vehicle configuration. Nonholonomic rolling constraints further shape the forces required for propulsion and turning. The controller must coordinate stepping, posture, and arm motion across starting, sustained pulling, turning, and stopping. During deployment, this coordination must be achieved without direct observations of vehicle state, interaction forces, or load parameters.

Recent advances in reinforcement learning have enabled robust sim-to-real locomotion~\cite{tan2018sim,hwangbo2019learning,lee2020learning,siekmann2021blind,radosavovic2024realworld,miki2022learning}. Whole-body control methods extend these capabilities to object transportation, sequential contacts, and sustained dynamic interaction~\cite{dao2024box,zhang2025wococo,sleiman2021unified,zhang2026falcon,xu2025facet}. Humanoid interaction with passive wheeled systems has also been explored through model-based loaded-cart transportation~\cite{rioux2015humanoid,rioux2018autonomous} and learning-based approaches to agile object pushing and pulling~\cite{li2026haic,li2026vaic}. Building on these developments, we focus on sustained rickshaw pulling across different load configurations. The robot must maintain stable, command-responsive locomotion while coordinating propulsion and steering through persistent handle contacts and adapting to load- and condition-dependent vehicle dynamics.

Randomizing physical parameters during training can improve policy robustness to dynamics variations and facilitate sim-to-real transfer~\cite{peng2018sim,tan2018sim}. In humanoid rickshaw pulling, similar instantaneous humanoid observations can correspond to different vehicle states and load parameters, and may require different control responses. Observation histories provide cues about the coupled dynamics by capturing the humanoid's response to previous actions. Prior work combines dynamics randomization with memory-based policies to support implicit dynamics inference~\cite{peng2018sim}, while legged-locomotion systems use proprioceptive histories for adaptation under varying physical conditions~\cite{lee2020learning,kumar2021rma,radosavovic2024realworld}. Privileged learning and asymmetric actor--critic training additionally exploit simulator-only information to guide policy learning~\cite{pinto2018asymmetric}. These complementary ideas motivate privileged supervision of a control-relevant representation of humanoid--vehicle interaction that can be inferred from onboard observation history during deployment.

We develop a whole-body controller using privileged teacher–student learning. The teacher uses vehicle states, interaction forces, and load properties to guide action and latent distillation into a student using onboard observation history. We then jointly fine-tune the student encoder and action head to refine control with the inferred representation. Training spans randomized vehicle, payload, and terrain conditions, while policy comparisons examine the roles of observation history, privileged training, and arm coordination.

Beyond reliable pulling, we investigate how persistent interaction with a wheeled load changes humanoid motion and actuation demands. Handle reactions influence the robot’s balance. We examine this interaction through gait-resolved handle forces, centroidal motion, and joint-effort redistribution. Positive mechanical work and squared-torque proxies characterize actuation demands across load--speed conditions, with walking comparisons identifying where pulling reduces the robot-normalized squared-torque proxy.

The main contributions of this work are:
\begin{itemize}
    \item A whole-body learning framework for humanoid rickshaw pulling that coordinates vehicle-motion tracking, balance, and persistent bimanual contact. Privileged teacher–student training transfers vehicle, interaction, and load information into a deployable controller using onboard observation history, without direct vehicle-state, contact-force, or load-parameter measurements.

    \item An empirical characterization of humanoid–rickshaw interaction that identifies load-dependent joint-effort redistribution and gait-synchronized handle reactions consistent with damping-like effects on lateral and roll motion. The analysis identifies conditions under which pulling yields a lower robot-normalized squared-torque proxy than the evaluated walking baseline.

    \item Real-world validation of a single policy with robot and human passengers at loaded-rickshaw masses of 60--115~kg. Without load-specific retuning, the policy executes starting, sustained pulling, and stopping, with the heaviest loaded rickshaw approximately 3.15 times the robot's mass.
\end{itemize}

\section{Related Work}
\label{sec:related_work}

\subsection{History-Based Locomotion Adaptation}

Observation--action history provides information about unobserved dynamics for locomotion adaptation~\cite{lee2020learning,radosavovic2024realworld}. RMA conditions a base policy on privileged environmental properties encoded as a latent, which an adaptation module estimates from history during deployment~\cite{kumar2021rma}. A-RMA subsequently fine-tunes the policy with the adaptation module fixed to accommodate estimation errors in bipedal locomotion~\cite{kumar2022arma}. History-conditioned transformers provide another route, learning action selection from observation--action sequences through reinforcement learning~\cite{radosavovic2024realworld}. These approaches distinguish the use of history from the training procedure that shapes its representation.

\subsection{Whole-Body Loco-Manipulation}

Whole-body loco-manipulation coordinates locomotion, balance, and object interaction through a common control objective. Unified MPC explicitly couples robot and object dynamics in multi-contact optimization~\cite{sleiman2021unified}. Learning-based methods realize box pickup and carrying on Digit~\cite{dao2024box}, while WoCoCo organizes diverse humanoid tasks through sequential contact stages~\cite{zhang2025wococo}. COLA learns proprioception-only human--humanoid collaborative carrying~\cite{du2026cola}. FALCON jointly trains lower-body locomotion and upper-body position control with adaptive force compensation, including real-world cart pulling~\cite{zhang2026falcon}. HMC combines position, impedance, and hybrid force--position control
for contact-rich humanoid manipulation~\cite{wei2026hmc}.These studies establish complementary approaches to coordinating body motion and contact forces.

\subsection{Humanoid Interaction with Passive Wheeled Systems}

Passive wheeled objects have independent motion constrained by wheel--ground contact, making their dynamics part of the interaction task. Rioux and Suleiman combine load-dependent motion primitives, whole-body control, and navigation for loaded-cart transport, including suppression of transmitted oscillations~\cite{rioux2018autonomous}. HUSKY exploits skateboard tilt--steering coupling to coordinate pushing, steering, and contact transitions~\cite{han2026husky}. HAIC infers object dynamics from proprioceptive history and demonstrates agile cart pushing and pulling under varying loads~\cite{li2026haic}. Building on these studies, we investigate sustained bimanual rickshaw pulling through simulation analyses and real-world evaluation across load configurations.

\section{Problem Formulation}
\label{sec:problem}
 
\begin{figure}[t]
\vspace*{1.5pt}
    \centering
    \includegraphics[width=\linewidth]{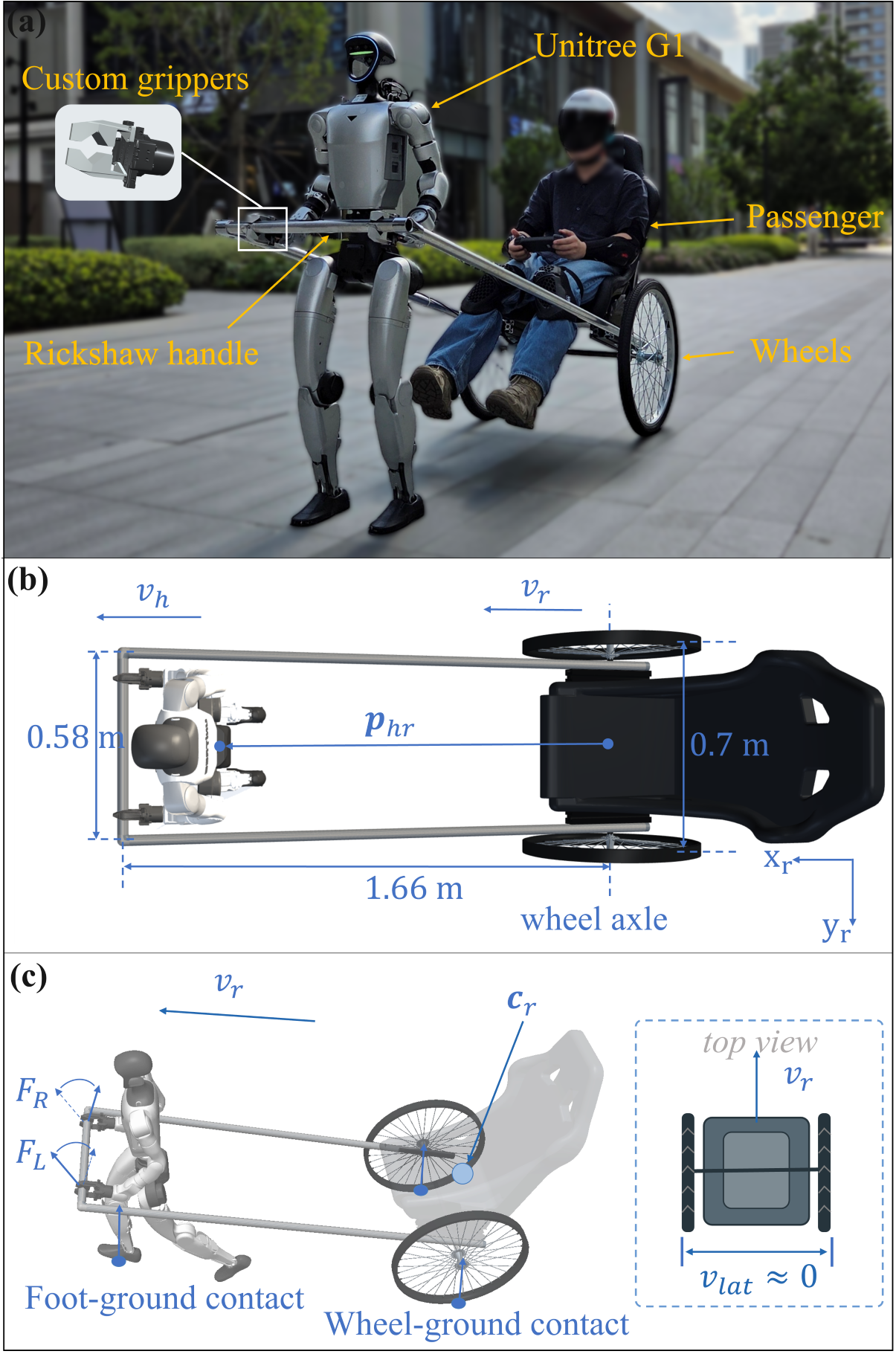}
\caption{\textbf{Humanoid--rickshaw system.} 
(a) The physical setup connects G1 to a passive rickshaw through custom
grippers. (b) Relative geometry characterizes the coupled configuration.
(c) G1 applies pulling forces 
%\(-\bm F_L,-\bm F_R\) 
through the handles, while the wheels support the load and constrain lateral motion.}
\label{fig:mechanical_design}
\end{figure}

We formulate a humanoid robot’s pulling of a passive two-wheeled rickshaw through persistent bimanual grasping as a partially observed MDP. The task is to realize desired vehicle velocity while maintaining balance and grasp, without directly observing vehicle states or load parameters during deployment. Fig. \ref{fig:mechanical_design} shows the hardware system and its kinematics.

\subsection{Robot System}

\block{Humanoid and Gripper}
We use a Unitree G1 with $29$ actuated joints spanning the legs, waist, and arms, and two Dex1-1 grippers on its hands. Its proprioceptive state contains the base orientation and angular velocity, joint positions, and joint velocities. The original Dex1-1 gripper is designed for table-top manipulation rather than handling rickshaw with heavy payload; therefore, we install custom-designed end-effectors (made of 6061 Aluminum) on the grippers.

\block{Rickshaw}
Because commercial rickshaws are too large for G1, we build a suitably sized vehicle using off-the-shelf components, including bicycle wheels, a racing bucket seat, and aluminum tubing and profiles. The unloaded rickshaw has a mass of 22.8 Kg. The rickshaw's unactuated wheels support a payload of variable mass and mass distribution, while handle wrenches by G1 provide propulsion and steering.

\block{Policy Action}
At 50~Hz, the policy outputs \(\bm a_t\in\mathbb R^{29}\), yielding
joint targets \(\bm q^{\mathrm{des}}_t=\bm q^0+\bm S_a\bm a_t\), where \(\bm q^0\in\mathbb R^{29}\) is the nominal joint-position reference and
\(\bm S_a\) is a diagonal scaling matrix. Joint PD controllers track
these targets; gripper PD controllers use fixed closing targets.

\block{Robot Observation}
The deployed policy receives onboard proprioception, the task command,
and the previous action:
\begin{equation}
\bm o_t=[\bm\omega_t,\ \bm g_t,\ \bm c_t,\
\bm q_t-\bm q^0, \dot{\bm q}_t,\ \dot{\bm q}_{g,t},\ \bm a_{t-1}]
\in\mathbb R^{97},
\label{eq:actor_observation}
\end{equation}
where \(\bm\omega_t,\bm g_t\) are base angular velocity and projected
gravity in the humanoid base frame, \(\bm q_t,\dot{\bm q}_t\in\mathbb R^{29}\) are joint positions and velocities, and \(\dot{\bm q}_{g,t}\in\mathbb R^2\) contains gripper velocities. Vehicle state, base linear velocity, contact forces, and randomized parameters are unavailable to the student actor.
 
\subsection{Markov Decision Process (MDP)}

\block{Task Command}
Vehicle motion is expressed in the rickshaw frame: its forward
axis follows the wheel rolling direction, its normal axis follows the
terrain normal, and its lateral axis completes a right-handed frame.
The command \(\bm c_t=[v_{r,t}^{\mathrm{cmd}},\omega_{r,t}^{\mathrm{cmd}}]^\top\)
specifies forward speed and yaw rate about the terrain normal. 
Lateral
velocity is uncommanded, consistent with the wheel constraint. Measured
forward speed, lateral velocity, and yaw rate are \(v_r,v_{\mathrm{lat}},
\omega_r\); \(\omega_{\mathrm{pitch}}\) is angular velocity about the
lateral axis.

\block{History and Objective}
The policy uses the current observation \(\bm o_t\) and the preceding \(H=61\) observations \(\bm o_{t-H:t-1}\).
We maximize 
% \begin{equation}
%     \max_{\pi}\;
%     \mathbb{E}_{\pi}
%     \left[
%         \sum_{t=0}^{T-1}\gamma^t r_t
%     \right], \textstyle
%     \label{eq:rl_objective}
% \end{equation}
\begin{equation}
\max_{\pi}\;
\mathbb{E}_{\pi}\!\left[\textstyle\sum_{t=0}^{T-1}
\gamma^t r_t\right].
\label{eq:rl_objective}
\end{equation}
where \(\gamma\) is the discount factor, \(T\) is the horizon, and \(r_t\) denotes the reward in Sec.~\ref{sec:method}.

% beam
% ==========

\begin{figure*}[!thp]
\vspace*{1.5pt}
    \centering
    \includegraphics[width=0.9\linewidth]{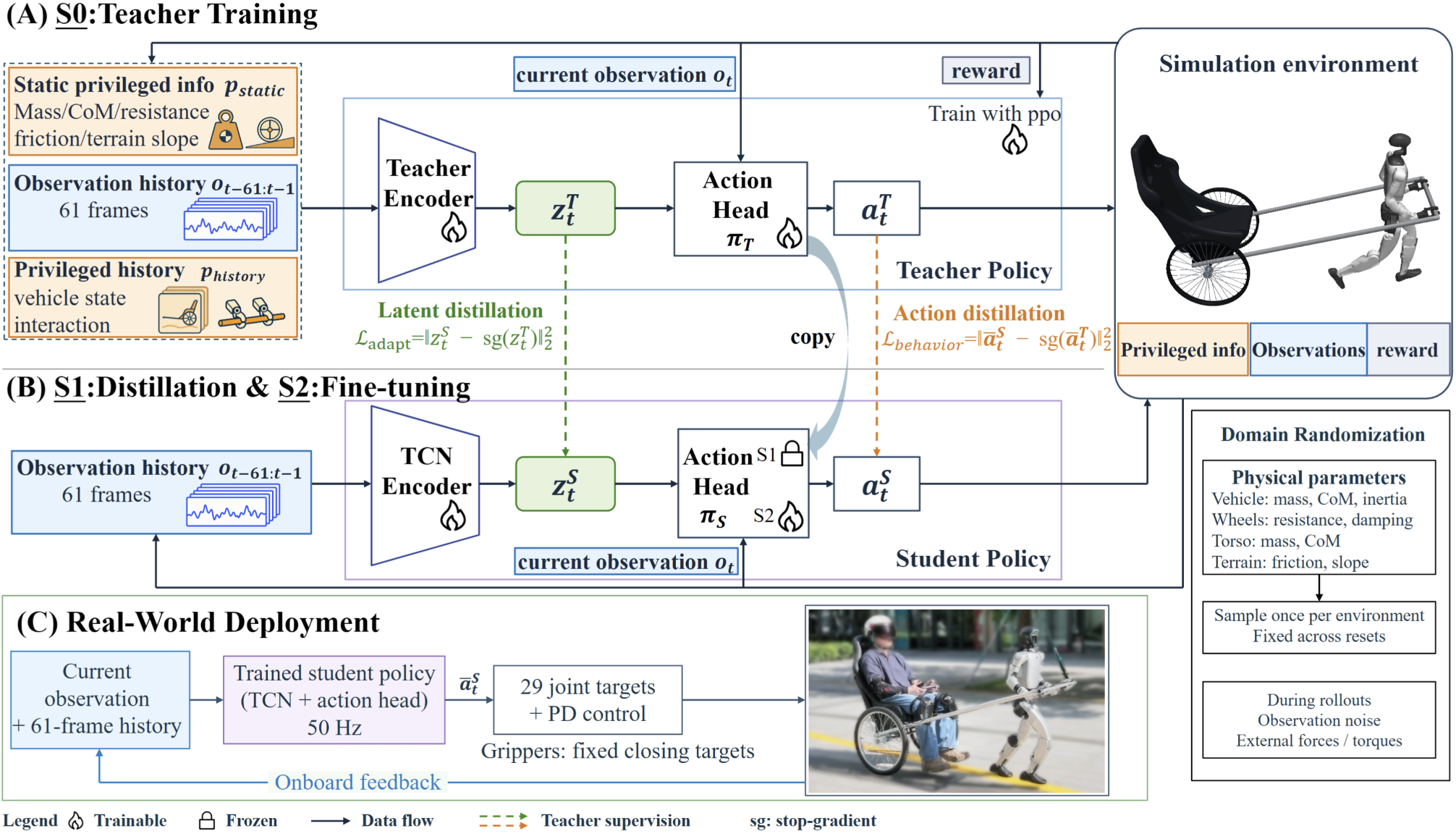}
\caption{\textbf{Teacher--student training and deployment.}
We first train a teacher with privileged load and interaction information, then distill its actions and latent into a student using only onboard observation history. Joint encoder and policy fine-tuning completes training for deployment without privileged inputs.}
\label{fig:approach}
\end{figure*}

\section{Method}
\label{sec:method}

Proprioceptive responses during pulling provide cues about the unobservable vehicle dynamics. To learn a control-relevant representation from these responses, we train a privileged teacher using robot and vehicle histories together with static parameters (\underline{S0}). Its actions and latent are distilled into a history-conditioned student (\underline{S1}). We then jointly fine-tune the student encoder and action head with PPO~\cite{schulman2017ppo} (\underline{S2}; Fig.~\ref{fig:approach}). This approach shares the motivation of prior history-based adaptation methods~\cite{kumar2021rma,kumar2022arma}.

\subsection{Privileged Teacher Policy}

\block{Privileged Information}
Handle forces drive the passive rickshaw while their reactions affect humanoid balance. The static input \(p_{\mathrm{static}}=[m_r,\bm c_r,c_{\mathrm{rr}},\mu,\alpha]\in\mathbb R^7\) describes the conditions governing this interaction: loaded-rickshaw mass, body-frame CoM, rolling resistance, ground friction, and terrain inclination.

Dynamic privileged histories \(p_{\mathrm{history}}=\bm d_{t-H:t-1}\in\mathbb R^{H\times17}\), together with robot observation histories, allow the teacher to learn how the vehicle responds to the humanoid’s actions, and use this to regulate the coupled system. Each frame contains
\begin{equation}
\bm d_t = [v_{r,t},\omega_{r,t},\theta_{r,t},
\bm F_{L,t},\bm F_{R,t},\bm p_{hr,t}, \psi_{hr,t},\psi_{tr,t},\bm v_{h,t}]\in\mathbb R^{17}.
 \nonumber
\label{eq:teacher_dynamic}
\end{equation}
Here, \(\theta_r\) is vehicle pitch; \(\bm F_L,\bm F_R\) are the forces on the left and right grippers. These forces and the humanoid base position relative to the rickshaw \(\bm p_{hr}\) use the rickshaw frame (Sec.~\ref{sec:problem}). \(\psi_{hr}\) and \(\psi_{tr}\) denote the pelvis and torso yaw angles relative to the rickshaw, respectively. The humanoid base velocity \(\bm v_h\) is expressed in the base frame.

\block{Teacher}
The teacher encoder combines a causal temporal convolutional network (TCN)~\cite{bai2018tcn} and a static branch:
\begin{equation}
\bm z^{\mathrm T}_t=f_{\mathrm T}(\bm o_{t-H:t-1},p_{\mathrm{history}},p_{\mathrm{static}})
 \in\mathbb R^{12},
 \label{eq:teacher_latent}
\end{equation}
where superscripts \(\mathrm T\) denote the teacher.

The TCN receives per-step concatenations of \(\bm o_\tau\) and \(\bm d_\tau\) through a pointwise \(114\!\rightarrow\!64\) projection and four 64-channel residual blocks with kernel size 5 and dilations \(1,2,4,8\). The resulting 61-step receptive field matches \(H\). Each block contains a dilated and a pointwise convolution, with ELU after the dilated convolution and residual addition. Left padding preserves causality and length. The last feature and temporal mean are concatenated into 128 dimensions.

A \(7\!\rightarrow\!16\) linear layer with ELU encodes \(p_{\mathrm{static}}\). Its output is concatenated with the temporal features and passed through a \(144\!\rightarrow\!64\!\rightarrow\!12\) MLP with an ELU hidden activation. Increasing the latent dimension from 12 to 24 showed no appreciable performance difference.

\block{Action Head}
Both policies map the current observation and latent \([\bm o_t,\bm z_t]\) through a 512--256--128 MLP with ELU hidden activations and a linear 29-D Gaussian mean output. \underline{S0} jointly trains the teacher encoder and action head with PPO and an asymmetric critic receiving additional vehicle, contact, and domain information.

\subsection{History-Based Student Adaptation}

\block{Adaptation Module}
The student (denoted by \(\mathrm S\)) uses onboard history alone:
\begin{equation}
 \bm z^{\mathrm S}_t=f_{\mathrm S}(\bm o_{t-H:t-1})\in\mathbb R^{12}.
 \label{eq:student_latent}
\end{equation}
Its encoder retains the TCN blocks and temporal aggregation, uses a \(97\!\rightarrow\!64\) input projection, and replaces static fusion with a \(128\!\rightarrow\!64\!\rightarrow\!12\) MLP with an ELU hidden activation. The policy is %$\bm a_t\sim\pi_{\mathrm S}(\,\cdot\mid\bm o_t,\bm z^{\mathrm S}_t)$.
\begin{equation}
 \bm a_t\sim\pi_{\mathrm S}(\,\cdot\mid\bm o_t,\bm z^{\mathrm S}_t).
 \label{eq:student_policy}
\end{equation}

\block{Action and Latent Distillation}
In \underline{S1}, the student copies and freezes the teacher's action head. Only \(f_{\mathrm S}\) is trained against the frozen teacher's targets on student-generated trajectories:
\begin{equation}
 \begin{aligned}
 \mathcal L_{\mathrm{S1}}
 &=\|\bar{\bm a}^{\mathrm S}_t-\bar{\bm a}^{\mathrm T}_t\|_2^2
 +\|\bm z^{\mathrm S}_t-\bm z^{\mathrm T}_t\|_2^2,
 \end{aligned}
 \label{eq:distillation}
\end{equation}
where \(\bar{\bm a}_t\) is the deterministic policy output.

\block{Policy Fine-Tuning}
After \underline{S1}, residual latent-estimation errors can induce action deviations from the teacher, impairing vehicle tracking and balance. In \underline{S2}, we unfreeze the action head and jointly fine-tune it with the student encoder using PPO under \eqref{eq:rl_objective}, adapting the policy to the estimated latent and refining the latent representation.

\subsection{Coupled Vehicle--Humanoid Reward Design}
\label{sec:reward_design}

Both PPO stages reward commanded vehicle motion while maintaining humanoid balance and grasp: $r_t=r_t^{\mathrm{task}}+r_t^{\mathrm{stability}}
+r_t^{\mathrm{interaction}}+r_t^{\mathrm{regularization}}$.
% \begin{equation}
% r_t=r_t^{\mathrm{task}}+r_t^{\mathrm{stability}}
% +r_t^{\mathrm{interaction}}+r_t^{\mathrm{regularization}}.
% \label{eq:reward_groups}
% \end{equation}
Speed and yaw-rate rewards track vehicle motion; posture and body-motion terms encourage balance. Interaction penalties regulate relative pose, grasp geometry, vehicle pitch, and pitch rate, while regularization discourages excessive actuation, abrupt actions, and unsafe configurations. Episodes terminate upon a fall or excessive gripper–handle separation to reduce unnecessary exploration.

For the feet, a minimum lateral spacing discourages narrow or crossed steps. The velocity-weighted clearance cost regulates height during horizontal motion, while the peak-height cost evaluates the completed swing at touchdown. Slip and landing-force penalties encourage traction and softer landings. Table~\ref{tab:reward_terms} lists all terms and weights.

\begin{table}[t]
\vspace*{1.5pt}
\centering
\caption{Whole-body control rewards and weights.}
\label{tab:reward_terms}
\scriptsize
\setlength{\tabcolsep}{1pt}
\renewcommand{\arraystretch}{1.04}
% Column widths plus the four internal padding spaces equal one column.
\begin{tabular}{@{}
>{\raggedright\arraybackslash}p{0.39\columnwidth}
>{\raggedright\arraybackslash}p{0.49\columnwidth}
>{\centering\arraybackslash}p{\dimexpr0.12\columnwidth-4\tabcolsep\relax}
@{}}
\toprule
Term & Expression & Weight \\
\midrule
\multicolumn{3}{l}{\textbf{Task tracking}}\\
Rickshaw forward speed
& \(\exp[-(v_r-v_r^{\mathrm{cmd}})^2/0.2]\) & \(2.0\) \\
Rickshaw yaw rate
& \(\exp[-(\omega_r-\omega_r^{\mathrm{cmd}})^2/0.1]\) & \(3.0\) \\
\midrule
\multicolumn{3}{l}{\textbf{Stability and gait}}\\
Torso upright
& \(\exp(-\|\bm g_{\mathrm{torso},xy}\|_2^2/0.2)\) & \(0.5\) \\
Posture
& \(\exp[-\frac{1}{29}\sum_j(\Delta q_j/\sigma_j)^2]\) & \(1.0\) \\
Torso angular velocity
& \(\|\bm\omega_{\mathrm{torso},xy}\|_2^2\) & \(-0.05\) \\
Whole-body angular momentum
& \(\|\bm L_h^C\|_2^2\) & \(-0.02\) \\
Minimum foot spacing
& \([(0.11-d_f)/0.11]_+^2\) & \(-1.0\) \\
Foot clearance
& \(I_{move}\sum_i|h_i-0.10|\|\bm v_{i,xy}\|_2\) & \(-2.0\) \\
Swing height at landing
& \(I_{move}\sum_i(h_i^{\mathrm{peak}}/0.10-1)^2
I_i^l\) & \(-0.25\) \\
Foot slip
& \(I_{move}\sum_i I_i^c
\|\bm v_{i,xy}\|_2^2\) & \(-0.1\) \\
Landing impact
& \(I_{move}\sum_i I_i^l\|\bm F_i\|_2\)
& \(-10^{-5}\) \\
\midrule
\multicolumn{3}{l}{\textbf{Rickshaw interaction}}\\
Humanoid--rickshaw position
& \(\Delta x^2+4\Delta y^2+\Delta z^2\) & \(-10.0\) \\
Gripper--handle position
& \(4\sum_k(\Delta d_k)^2\) & \(-10.0\) \\
Relative yaw
& \(\frac12(\Delta\psi_{hr}^2+\Delta\psi_{tr}^2)\)
& \(-1.0\) \\
Rickshaw pitch
& \((\Delta\theta_r)^2\) & \(-5.0\) \\
Rickshaw pitch rate
& \(\omega_{\mathrm{pitch}}^2\) & \(-1.0\) \\
\midrule
\multicolumn{3}{l}{\textbf{Regularization and safety}}\\
Joint position limits
& \(\begin{aligned}[t]
  &\sum_j\bigl([q_j^{\min}-q_j]_++[q_j-q_j^{\max}]_+\bigr)
  \end{aligned}\) & \(-1.0\) \\
Effort-margin violation
& \(\sum_j\left[\frac{|\tau_j|/\tau_j^{\max}-0.75}{0.25}\right]_+^2\)
& \(-1.0\) \\
Positive mechanical power
& \(\sum_j[\tau_j\dot q_j]_+\) & \(-10^{-3}\) \\
Action rate
& \(\|\bm a_t-\bm a_{t-1}\|_2^2\) & \(-0.1\) \\
Undesired collision
& \(N_{\mathrm{col}}\) & \(-1.0\) \\
\bottomrule
\end{tabular}
\par\vspace{2pt}
\begin{minipage}{\columnwidth}
\scriptsize\raggedright
Time indices are omitted. Negative weights multiply nonnegative costs; \(\Delta\) denotes deviation from a nominal reference, angular errors are wrapped to \([-\pi,\pi]\), and \([x]_+=\max(x,0)\). Position errors satisfy \((\Delta x,\Delta y,\Delta z)^\top=\bm p_{hr}-\bm p_{hr}^0\); \(d_k\) is a gripper--handle distance. Both relative-yaw errors use the initial rickshaw--pelvis heading difference \(\psi_{hr}^0\). For the foot terms, \(d_f\) is the left-minus-right lateral spacing in the humanoid base frame, and \(h_i^{\mathrm{peak}}\) is peak swing height. Contact and landing indicators are \(I_i^c,I_i^l\); the motion gate is \(I_{move}=\mathbf{1}\{|v_r^{\mathrm{cmd}}|+|\omega_r^{\mathrm{cmd}}|>0.1\}\).
\end{minipage}
\end{table}

\subsection{Training and Deployment}

\block{Training Distribution}
We randomize physical parameters, observations, and external
disturbances as specified in Table~\ref{tab:domain_randomization}.
Physical parameters are sampled once per environment and held
fixed across resets.
Training commands span
\(v_r^{\mathrm{cmd}}\in[-0.6,2.0]\)~m/s and
\(\omega_r^{\mathrm{cmd}}\in[-0.4,0.4]\)~rad/s. Higher-speed pulling is achieved through additional policy fine-tuning (Fig.~\ref{fig:rickshaw_overview}).

\block{Training and Deployment}
We train the three-stage pipeline in MuJoCo-based
mjlab~\cite{todorov2012mujoco,zakka2026mjlab} using 8192 parallel
environments on two NVIDIA H200 GPUs, requiring approximately
11~h.
The critic shares the action head's hidden widths and activations
and outputs a scalar value.
Actor and critic inputs use empirical observation normalization.
At deployment, the student uses the deterministic action mean,
converted to joint targets as described in Sec.~\ref{sec:problem}.

\begin{table}[t]
\vspace*{1.5pt}
\centering
\caption{Domain-randomization ranges.}
\label{tab:domain_randomization}
\scriptsize
\setlength{\tabcolsep}{2.5pt}
\renewcommand{\arraystretch}{1.05}
\begin{tabular}{@{}
>{\raggedright\arraybackslash}p{0.40\columnwidth}
>{\raggedright\arraybackslash}p{0.37\columnwidth}
>{\centering\arraybackslash}p{\dimexpr0.23\columnwidth-4\tabcolsep\relax}
@{}}
\toprule
Parameter & Range & Unit \\
\midrule
Rickshaw total mass & \(\mathcal U(20,60)\) & kg \\
Rickshaw CoM offset & \(\mathcal U(-0.10,0.10)\) & m \\
Rickshaw inertia scale & \(\mathcal U(0.8,1.2)\) & -- \\
Rolling resistance & \(\mathcal U(0.01,0.03)\) & -- \\
Wheel damping & \(\mathcal U(0.01,0.03)\) & \(\mathrm{Nm\,s/rad}\) \\
Torso mass offset & \(\mathcal U(-2,2)\) & kg \\
Torso CoM offset & \(\mathcal U(-0.05,0.05)\) & m \\
Ground friction & \(\mathcal U(0.3,1.2)\) & -- \\
Terrain slope & \(\{-0.05{:}0.01{:}0.05\}\) & rad \\
Observation noise
& \(\pm(0.04,0.05,0.01,\)\newline
  \(\phantom{\pm(}0.075,0.046)\) & Scaled \\
Robot force / torque & \(\pm60\,/\,\pm8\) & N / Nm \\
Rickshaw force / torque & \(\pm40\,/\,\pm6\) & N / Nm \\
Disturbance duration / cooldown & 0.05--0.15 / 1--3 & s \\
\bottomrule
\end{tabular}
\par\vspace{2pt}
\begin{minipage}{\columnwidth}
\scriptsize\raggedright
\(\mathcal U\) denotes a uniform distribution. CoM offsets and inertia scales apply independently per axis. Observation-noise bounds are given to \((\bm\omega,\bm g,\bm q,\dot{\bm q},\dot{\bm q}_g)\).
\end{minipage}
\end{table}

\section{Experimental Results}
\label{sec:results}

We evaluate vehicle tracking across loads and the contribution of each design choice. Beyond tracking, joint-effort and handle-interaction analyses help explain how pulling a wheeled load changes locomotion, transport costs, and balance behavior. Hardware trials evaluate different passenger configurations, revealing the remaining sim-to-real differences.

\subsection{Experimental Setup and Metrics}

We compare the \emph{Privileged Teacher}, the distilled and fine-tuned student (\emph{Ours}), a policy trained without privileged distillation (\emph{Only History}), a current-observation policy (\emph{No History}), and a \emph{Fixed Arm} variant that follows \emph{Ours} but holds the arm joint targets \(\mathbf q_{\mathrm{arm}}^{\mathrm{des}}\) constant. Each variant is trained with seeds 0--3 and evaluated using the training DR range (Table~\ref{tab:domain_randomization}). To isolate load-dependent behavior, we separately sweep rickshaw mass over 20--120~kg and speed commands over 0.6--2.0~m/s on flat terrain, with other parameters fixed at nominal values. Masses above 60~kg exceed the training range. A G1 walking policy trained with the default mjlab settings serves as the baseline for energy-consumption and locomotion-behavior comparisons.

Table~\ref{tab:policy_comparison} reports tracking RMSE, vehicle motion, torso tilt, and a cost-of-transport (CoT) proxy~\cite{wang2026skating}. We extend the robot-mass normalization to include the rickshaw for system-level comparisons:
\begin{equation}
\begin{aligned}
 \mathrm{CoT}_{\mathrm{robot}}
 &=\frac{\|\bm\tau\|_2^2}
 {m_h g\|\bm\xi\|_2}, 
 \mathrm{CoT}_{\mathrm{system}}
 =\frac{m_h}{m_h+m_r}\mathrm{CoT}_{\mathrm{robot}}. \textstyle
\end{aligned} \nonumber
% \label{eq:normalized_torque_cost}
\end{equation}
Here, \(m_h,m_r\) are robot and total rickshaw masses, and \(\bm\xi\) is the actual planar twist of the rickshaw during pulling or the robot during walking. This squared-torque proxy measures actuation effort.

We also report mechanical cost of transport (MCOT), counting positive actuator work as in Griffin and Grizzle~\cite{griffin2017nonholonomic}: $\mathrm{MCOT}
 =\frac{1}{m_h g s_r}\int_0^{T_{eval}}\sum_j\max\!\left(\tau_j(t)\dot q_j(t),0\right)\,\mathrm{d}t$,
where \(T_{eval}\) is the evaluation duration and \(s_r\) is the corresponding rickshaw travel distance. 

\subsection{Vehicle Tracking and Policy Ablations}

As shown in Table~\ref{tab:policy_comparison}, \emph{Ours}  achieves forward-speed and yaw-rate RMSEs of 0.291~m/s and 0.0535~rad/s. The \emph{Privileged Teacher} retains lower tracking errors and pitch-rate RMS. \emph{Only History} reduces forward and yaw tracking RMSE by 23.3\% and 34.8\% relative to \emph{No History}. At similar tracking errors, \emph{Ours} further reduces pitch-rate RMS, yaw-acceleration RMS, torso-tilt RMS, and $\mathrm{CoT}_{\mathrm{system}}$ relative to \emph{Only History} by 2.9\%, 4.1\%, 4.9\%, and 4.6\%, respectively. Compared with \emph{Fixed Arm}, \emph{Ours} reduces forward RMSE by 52.3\% and pitch-rate RMS by 31.6\%, supporting coordinated arm control.

\begin{figure}[t]
\vspace*{1.5pt}
    \centering
    \includegraphics[width=\columnwidth]{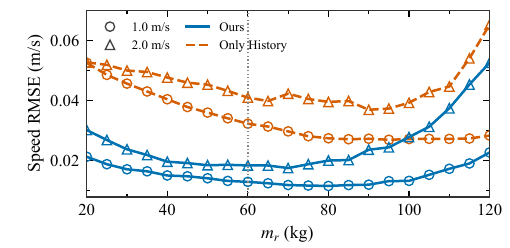}
    \caption{\textbf{Vehicle tracking and load generalization.}
    \emph{Ours} and \emph{Only History} are compared during constant-speed pulling across loads within and beyond the training range.}
    \label{fig:tracking_performance}
\end{figure}

Fig.~\ref{fig:tracking_performance} shows that \emph{Ours} generalizes to masses beyond the training range and outperforms \emph{Only History} across the tested loads. The advantage is particularly pronounced within the mass DR range, consistent with the teacher's access to static load parameters providing useful guidance for adaptation.

\begin{table}[t]
\vspace*{1.5pt}
\centering
\caption{Policy comparison.} % under training-distribution conditions.}
\label{tab:policy_comparison}
\scriptsize
\setlength{\tabcolsep}{3.3pt}
\renewcommand{\arraystretch}{1.12}
\begin{tabular}{@{}lccccc@{}}
\toprule
Metric
& \shortstack{Privileged\\Teacher}
& \textbf{Ours}
& \shortstack{Only\\History}
& \shortstack{No\\History}
& \shortstack{Fixed\\Arm} \\
\midrule
\textbf{Tracking} \\
Forward RMSE (m/s)
& 0.278 & \textbf{0.291} & 0.292 & 0.381 & 0.609 \\

Yaw-rate RMSE (rad/s)
& 0.0450 & 0.0535 & \textbf{0.0528} & 0.0809 & 0.0703 \\
\midrule
\textbf{Rickshaw Motion} \\
Pitch-rate RMS (rad/s)
& 0.0481 & \textbf{0.0530} & 0.0546 & 0.0646 & 0.0775 \\

Forward-accel.\ RMS (m/s$^2$)
& 0.995 & \textbf{0.960} & 0.975 & 1.205 & 1.154 \\

Yaw-accel.\ RMS (rad/s$^2$)
& 0.763 & \textbf{0.764} & 0.797 & 1.204 & 0.902 \\
\midrule
Torso-tilt RMS (rad)
& 0.0668 & \textbf{0.0651} & 0.0684 & 0.0789 & 0.0799 \\
\midrule
$\mathrm{CoT}_{\mathrm{system}}$ proxy
& 5.250 & \textbf{4.936} & 5.174 & 6.815 & 7.706 \\
\bottomrule
\end{tabular}
\par\vspace{2pt}
\begin{minipage}{\columnwidth}
% \scriptsize\raggedright
% Bold indicates the best non-privileged result.
\end{minipage}
\end{table}

\subsection{Transport Cost and Joint-Effort Redistribution}

\begin{figure*}[t]
\vspace*{1.5pt}
    \centering
    \includegraphics[width=\textwidth]{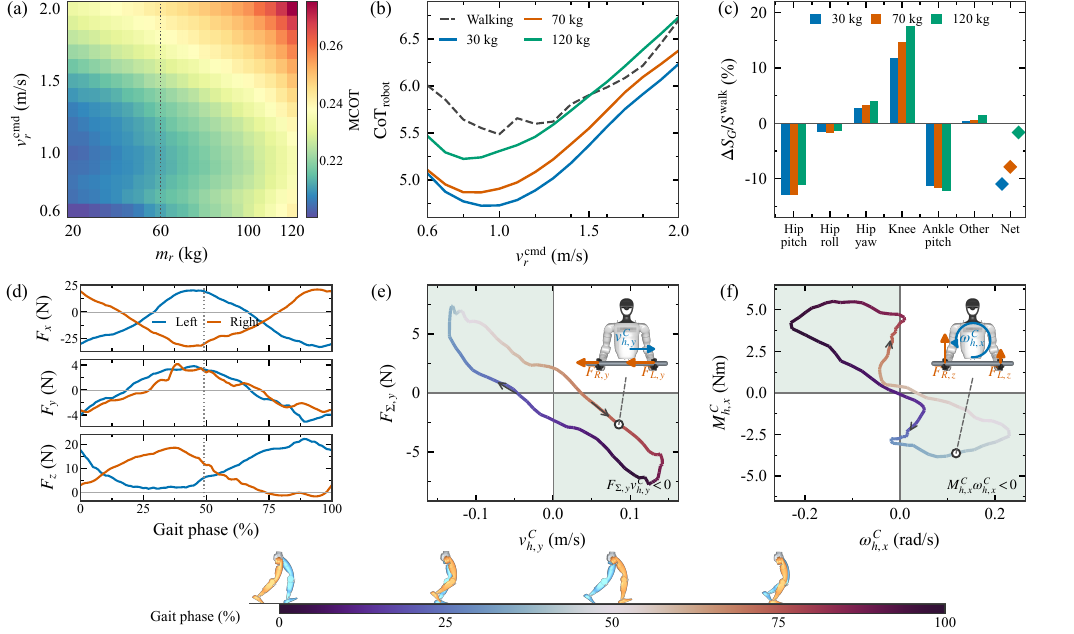}
    \caption{\textbf{Transport cost and contact coordination.}
   (a--c) Load--speed trends, comparison with walking, and joint-effort redistribution characterize the actuation demands of constant-speed pulling. (d--f) Gait-synchronized hand forces and their relation to centroidal motion illustrate the role of handle interaction in whole-body motion regulation.}
    \label{fig:transport_behavior}
\end{figure*}

MCOT generally increases with rickshaw mass and speed (Fig.~\ref{fig:transport_behavior}(a)). At 1.5~m/s, increasing mass from 30 to 120~kg raises MCOT from 0.214 to 0.250, a 16.5\% increase for four times the transported mass.

Despite towing additional mass, rickshaw pulling achieves lower \(\mathrm{CoT}_{\mathrm{robot}}\) than walking during constant-speed motion across most tested conditions (Fig.~\ref{fig:transport_behavior}(b)). Only at the heaviest load of 120~kg and higher speeds (1.6--2.0~m/s) does this actuation-cost proxy exceed the walking baseline.

Joint-effort redistribution helps explain the \(\mathrm{CoT}_{\mathrm{robot}}\) advantage (Fig.~\ref{fig:transport_behavior}(c)). We normalize each group's change \(\Delta S_G=S_G^{\mathrm{pull}}-S_G^{\mathrm{walk}}\), with \(S_G=\langle\sum_{j\in G}\tau_j^2\rangle\), by the walking total \(S^{\mathrm{walk}}\). At 70~kg, hip-pitch and ankle-pitch squared torques fall by 33\% and 61\%, outweighing a 126\% increase at the knees. Higher loads offset more of these savings, reducing the total squared-torque benefit from 10.9\% at 30~kg to 1.6\% at 120~kg.

\subsection{Gait-Synchronized Contact Coordination}

Through persistent grasp, G1 uses the rickshaw’s reaction forces to assist balance while towing it. This bidirectional interaction changes joint force and moment demands, contributing to the observed redistribution of joint effort. We examine how these reactions are coordinated over the gait cycle, and contribute to robot balance. Figure~\ref{fig:transport_behavior}(d) shows mean cycles at 70~kg and 1.5~m/s, with phase zero at left-foot landing. Forces act on G1 and are expressed in the rickshaw frame.

The lateral hand forces are approximately in phase, reinforcing the resultant \(F_{\Sigma,y}=F_{L,y}+F_{R,y}\). The longitudinal and vertical oscillations are nearly opposed, partially cancelling in the resultant while producing differential forces. Through the lateral handle spacing, vertical and longitudinal force differences contribute roll and yaw moments, respectively. This phase organization gives the contacts distinct roles in lateral and rotational motion; we examine their balancing contributions through the force--motion relationships in Fig.~\ref{fig:transport_behavior}(e,f).

For lateral motion, Fig.~\ref{fig:transport_behavior}(e) shows \(F_{\Sigma,y}v_{h,y}^{C}<0\) over most of the mean gait cycle, with a mean within-trial correlation of approximately \(-0.91\). The resultant reaction reverses with the gait and generally opposes CoM motion, providing a damping-like contribution to lateral balance.

For roll, Fig.~\ref{fig:transport_behavior}(f) similarly shows \(M^C_{h,x}\omega^C_{h,x}<0\) over most of the mean gait cycle: the rickshaw reaction moment about G1's CoM generally opposes its roll motion, contributing to roll balance. Here, \(\bm\omega_h^C=(\bm I_h^C)^{-1}\bm L_h^C\) is the average centroidal angular velocity~\cite{orin2013centroidal,saccon2017centroidal}, defined by the locked inertia and angular momentum about G1's CoM. Both moment and angular velocity use the rickshaw frame.

\subsection{Real-World Rickshaw Pulling}

\begin{figure*}[t]
\vspace*{1.5pt}
    \centering
    \includegraphics[width=\textwidth]{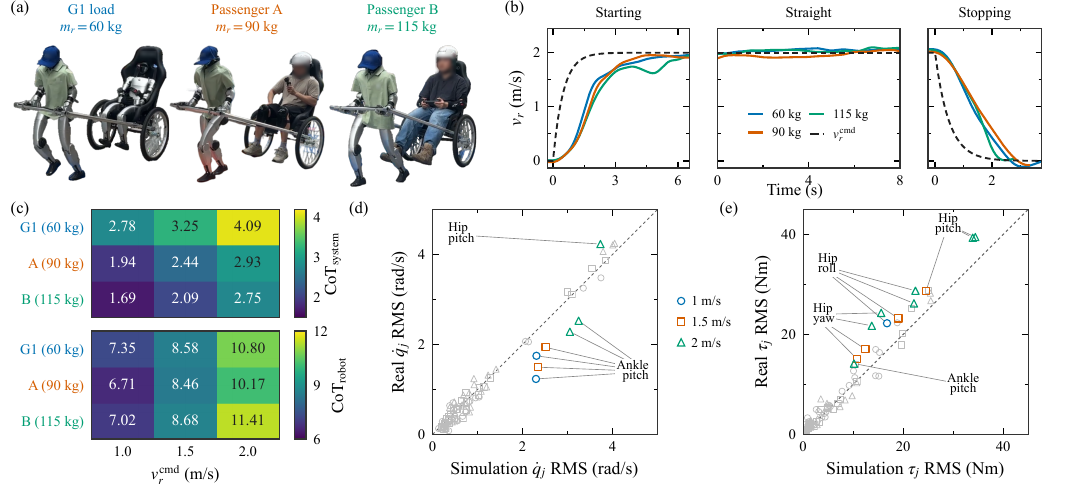}
    \caption{\textbf{Real-world deployment and sim-to-real comparison.} (a) Load scenarios tested with the same policy. (b) Starting, straight pulling, and stopping responses. (c) Robot- and system-normalized CoT. (d,e) Joint angular-velocity and torque RMS comparisons across three speeds for the seated-G1 scenario; labels identify notable discrepancies.}
    \label{fig:real_world}
\end{figure*}

We validate one fixed policy in three real-world scenarios: a seated G1 and two human passengers, with loaded-rickshaw masses of 60, 90, and 115~kg (Fig.~\ref{fig:real_world}(a)). The scenarios differ in mass, CoM, inertia, and passenger responses to vehicle motion, including posture adjustments. Each is tested at 1.0, 1.5, and 2.0~m/s. Vehicle speed is estimated from wheel-mounted IMU.

Across all scenarios, the rickshaw completes acceleration, constant-speed tracking, and stopping. Figure~\ref{fig:real_world}(b) illustrates these phases at 2.0 m/s, including the 115 kg load, approximately 3.15 times the robot’s mass.

Across the nine constant-speed pulling conditions (Fig.~\ref{fig:real_world}(c)), \(\mathrm{CoT}_{\mathrm{system}}\) ranges from 1.69 to 4.09 and \(\mathrm{CoT}_{\mathrm{robot}}\) from 6.71 to 11.41; both increase with speed within each scenario. At a fixed speed, the maximum \(\mathrm{CoT}_{\mathrm{robot}}\) is at most 12.2\% above the minimum despite loaded masses spanning 60--115~kg.  These results indicate favorable energy efficiency for heavy-load transport under the adopted CoT proxy.

Fig.~\ref{fig:real_world}(d,e) compares joint angular-velocity and torque RMS between simulation and hardware for the same policy in the seated-G1 scenario. Most joints show close agreement, with the main discrepancies occurring at the ankles and hips: hardware exhibits lower ankle-pitch angular-velocity RMS and higher hip torque RMS. These discrepancies may reflect differences between the simulated and physical systems in actuator dynamics and foot–ground contact behavior.

%%%%%%%%%%%%%%%%%%%%%%%%%%%%%%%%%%%%%%%%%%%%%%%%%%%%%%%%%%%%%%%%%%%%
\section{Conclusion and Future Work}
\label{sec:conclusion}
We presented a teacher--student whole-body controller for humanoid rickshaw pulling using onboard observation history. Privileged training improves sustained vehicle tracking over direct history-based learning, including loads beyond the training mass range. The behavior analysis shows that G1 both propels the rickshaw and uses its reactions to assist lateral and roll balance. This interaction accompanies joint-effort redistribution and a lower \(\mathrm{CoT}_{\mathrm{robot}}\) than walking over most of the tested load--speed conditions. The same policy accommodates distinct real-world dynamics with robot and human passenger loads, without load-specific retuning. Future work will address grasp acquisition and recovery, terrain-aware navigation, and electrical energy measurements in matched hardware walking and pulling trials.

\balance
% \addtolength{\textheight}{-3.7cm}   % This command serves to balance the column lengths
%                                   % the textheight of the last page by a suitable amount.
%                                   % This command does not take effect until the next page
%                                   % so it should come on the page before the last. Make
%                                   % sure that you do not shorten the textheight too much.

%%%%%%%%%%%%%%%%%%%%%%%%%%%%%%%%%%%%%%%%%%%%%%%%%%%%%%%%%%%%%%%%%%%%%%%%%%%%%%%%

%%%%%%%%%%%%%%%%%%%%%%%%%%%%%%%%%%%%%%%%%%%%%%%%%%%%%%%%%%%%%%%%%%%%%%%%%%%%%%%%

%%%%%%%%%%%%%%%%%%%%%%%%%%%%%%%%%%%%%%%%%%%%%%%%%%%%%%%%%%%%%%%%%%%%%%%%%%%%%%%%
% \begin{comment}


%     \section*{ACKNOWLEDGMENT}
% The authors would like to thank Sequoyah Walters and Yifen Chen for inspirational dicussions, Heidi Fehr for advising us on Mocap system, and Peter Adamcyzk for letting us use his lab space for doing the experiments.

% \end{comment}

%%%%%%%%%%%%%%%%%%%%%%%%%%%%%%%%%%%%%%%%%%%%%%%%%%%%%%%%%%%%%%%%%%%%%%%%%%%%%%%%
%  \newpage
% \bibliographystyle{IEEEtran}
% \bibliography{reference}

\section{Acknowledgement}
We would like to thank Rui Song for initial dicussion on the framework, and thank Hongzheng Zhu and Kunzhao Ren for their help on the experiments.

\bibliographystyle{IEEEtran}
\bibliography{ref/references,ref/IEEEcontrol}

\end{document}